\documentclass[letterpaper]{article} 
\usepackage{aaai19}  
\usepackage{times}  
\usepackage{helvet}  
\usepackage{courier}  
\usepackage{url}  
\usepackage{graphicx}  
\usepackage{flushend}
\def\year{2019}\relax

\nocopyright 

\usepackage{latexsym}
\usepackage{amsfonts}
\usepackage{amsmath}
\usepackage{amssymb}
\usepackage{booktabs}
\usepackage{caption}
\usepackage{color, colortbl}
\usepackage{comment}
\usepackage{courier}
\usepackage{empheq}
\usepackage{enumitem}
\usepackage{epstopdf}
\usepackage{graphicx}
\usepackage{hyperref}
\usepackage{listings}
\usepackage{makecell}
\usepackage{mathtools}
\usepackage{mdframed}
\usepackage{multirow}
\usepackage{natbib}
\usepackage{pbox}
\usepackage{qtree}
\usepackage{setspace}
\usepackage{stfloats}
\usepackage{subcaption}
\usepackage{tikz}
\usepackage{url}
\usepackage{verbatim}
\usepackage{xspace}

\usepackage[T1]{fontenc} 
\newcommand{\rom}[1]{\uppercase\expandafter{\romannumeral #1\relax}}
\newcommand{\eat}[1]{}
\usepackage{color,soul}
\usepackage{algorithm}
\usepackage{algpseudocode}
\algtext*{EndWhile}
\algtext*{EndIf}
\algtext*{EndFor}
\algtext*{EndProcedure}
\algtext*{EndFunction}

\newcommand{\LCA}{\mathrm{LCA}}

\newcommand{\Tanh}{\mathrm{Tanh}}

\title{A Transfer-Learnable Natural Language Interface for Databases}

\author{
  Wenlu Wang$^1$ \quad 
  Yingtao Tian$^2$ \quad
  Hongyu Xiong$^3$ \quad
  Haixun Wang$^4$ \quad
  Wei-Shinn Ku$^1$ 
  \vspace{1.6mm}\\
\fontsize{10}{10}\selectfont\itshape 
$^{1}$ Auburn University, Auburn, AL   \quad
\fontsize{10}{10}\selectfont\itshape 
$^{2}$ Stony Brook University, Stony Brook, NY\\
\fontsize{9}{9}\selectfont\itshape 
$^{3}$ Stanford University, Stanford, CA \quad
\fontsize{10}{10}\selectfont\itshape 
$^{4}$ WeWork Research   \\
\fontsize{9}{9}\selectfont\ttfamily\upshape
wenluwang@auburn.edu \quad yittian@cs.stonybrook.edu \quad hxiong2@stanford.edu\\
\fontsize{9}{9}\selectfont\ttfamily\upshape
haixun.wang@wework.com \quad weishinn@auburn.edu\\
  \\} 

\date{}

\begin{document}
\maketitle
\begin{abstract}

Relational database management systems (RDBMSs) are powerful because they are able to optimize and answer queries against any relational database. A natural language interface (NLI) for a database, on the other hand, is tailored to support that specific database. In this work, we introduce a general purpose transfer-learnable NLI with the goal of learning one model that can be used as NLI for any relational database. We adopt the data management principle of separating data and its schema, but with the additional support for the idiosyncrasy and complexity of natural languages. Specifically, we introduce an automatic annotation mechanism that separates the schema and the data, where the schema also covers knowledge about natural language. Furthermore, we propose a customized sequence model that translates annotated natural language queries to SQL statements. We show in experiments that our approach outperforms previous NLI methods on the WikiSQL dataset and the model we learned can be applied to another benchmark dataset OVERNIGHT without retraining.


\end{abstract}

\section{Introduction}



\begin{figure*}[b]
\vspace{-10pt}
\begin{subfigure}{.55\textwidth}
\scriptsize
\begin{center}\textbf{ \qquad Example (a)}\end{center}
\begin{tabular}{c|l}
\hline
 \rule{0pt}{8pt} \scriptsize \textbf{Question \textit{q}} &
Which \colorbox{olive!20}{film} \colorbox{blue!15}{directed by} \colorbox{orange!50}{Jerzy Antczak} did \colorbox{green!20}{Piotr Adamczyk} \colorbox{red!30}{star in}? \\
 \rule{0pt}{9.2pt} &  \\
\hline
 \rule{0pt}{8pt} \scriptsize \textbf{SQL \textit{s}} & 
 \scriptsize \texttt{SELECT \colorbox{olive!20}{Film\_Name} WHERE} \colorbox{blue!15}{Director} =\colorbox{orange!50}{``Jerzy Antcza''} \\
 \rule{0pt}{8pt} & \scriptsize \texttt{AND \colorbox{red!30}{Actor} = \colorbox{green!20}{``Piotr Adamczy''}}\\
\hline
 \rule{0pt}{8pt} \scriptsize  \textbf{\textit{Annotated}} &
 Which \colorbox{olive!20}{$c_1$ [film]}  \colorbox{blue!15}{$c_2$ [directed by]} \colorbox{orange!50}{$v_2$ [Jerzy Antczak]}did \colorbox{green!20}{$v_3$}\\
 \rule{0pt}{8pt} \scriptsize \textbf{\textit{Question q$_a$}} & \colorbox{green!20}{[Piotr Adamczyk]} \colorbox{red!30}{$c_3$ [Actor] star in} ?\\
\hline
\rule{0pt}{8pt} \scriptsize \textbf{\textit{Annotated}}& \multirow{2}{*}{\texttt{SELECT \colorbox{olive!20}{$c_1$} WHERE \colorbox{blue!15}{$c_2$} = \colorbox{orange!50}{$v_2$} AND \colorbox{red!30}{$c_3$} = \colorbox{green!20}{$v_3$}}}\\
\rule{0pt}{8pt} \textbf{\textit{SQL s$_a$}}& \\
\hline
\end{tabular}
\begin{flushright}
\resizebox{.92\textwidth}{!}{
\begin{tabular}{|c|c|c|c|c|}
\hline
\rowcolor{lightgray}
Nomination & \colorbox{red!30}{Actor} & \colorbox{olive!20}{Film\_Name} & \colorbox{blue!15}{Director} & Nomination Date\\
\hline
Best Actor in a Leading Role & \colorbox{green!20}{Piotr Adamczyk} & Chopin: Desire for Love & \colorbox{orange!50}{Jerzy Antczak} & 2003 August\\
\hline
Best Actor in a Supporting Role & Levan Uchaneishvili & 27 Stolen Kisses & Nana Djordjadze & 2003 August\\
\hline
\dots & \dots  & \dots & \dots & \dots \\
\hline
\end{tabular}
}
\end{flushright}
\end{subfigure}%
\hfill
\begin{subfigure}{.44\textwidth}
\scriptsize
\begin{center}\textbf{Example (b)}\end{center}
\begin{tabular}{|l}
\hline 
\rule{0pt}{8pt} \colorbox{olive!20}{How many people live in} \colorbox{orange!50}{Mayo} which has the \colorbox{red!30}{English name} \\
 \rule{0pt}{8pt}  \colorbox{green!20}{Carrowteige}? \\
\hline
 \rule{0pt}{8pt} \scriptsize 
\texttt{SELECT \colorbox{olive!20}{population} WHERE \colorbox{blue!15}{County} = \colorbox{orange!50}{``Mayo''} AND }\\ 
 \rule{0pt}{8pt} \texttt{\colorbox{red!30}{English\_Name} =\colorbox{green!20}{``Carrowteig''}}\\
\hline
 \rule{0pt}{8pt} \scriptsize  
\colorbox{olive!20}{$c_1$ [How many people live in]} 
	\colorbox{orange!50}{$v_2$ [Mayo]} 
	which has the 
 \colorbox{red!30}{$c_3$} \\
 \rule{0pt}{8pt}  \colorbox{red!30}{[English Name]} \colorbox{green!20}{$v_3$ [Carrowteige]} ?\\
    \hline
\rule{0pt}{8pt} \scriptsize 
	\multirow{2}{*}{\texttt{SELECT \colorbox{olive!20}{$c_1$} WHERE \colorbox{blue!15}{$c_2$} = \colorbox{orange!50}{$v_2$} AND 
		\colorbox{red!30}{$c_3$} = \colorbox{green!20}{$v_3$}}}\\
\rule{0pt}{8pt} \\
\hline
\end{tabular}
\begin{center}
\resizebox{.8\textwidth}{!}{
\begin{tabular}{|c|c|c|c|c|}
\hline
\rowcolor{lightgray}
\colorbox{blue!15}{County} & \colorbox{red!30}{English\_Name} & Irish\_Name & \colorbox{olive!20}{Population} & Irish\_Speakers\\
\hline
\colorbox{orange!50}{Mayo} & \colorbox{green!20}{Carrowteige} & Ceathru Thaidhg & 356 & 64\% \\
\hline
Galway & Aran Islands & Oileain Arann & 1225 & 79\%\\
\hline
\dots & \dots  & \dots & \dots & \dots \\
\hline
\end{tabular}
}
\end{center}
\end{subfigure}
\normalsize
\vspace{-5pt}
\caption{Natural language questions and their corresponding SQLs against two different databases. Note that the annotated SQLs of the two different questions are the same. This figure is better viewed on media with color support.}
\label{tbl:nlidb-example-1}
\end{figure*}

The majority of business data is relational data. Many applications are built on relational databases, including customer relations management~\cite{ngai2009application}, financial fraud detection~\cite{ngai2011application},  and knowledge discovery in medicine~\cite{esfandiari2014knowledge}, etc. However, a minimal understanding of SQL is required to operate on these data. This gives rise to the study of natural language interfaces to database~\cite{androu1995natural}, or NLIDB, with the goal of making databases more accessible to the general public. 

An NLIDB translates a natural language question to a structured query (in our case an SQL query) that can be executed by a database engine.
Figure~\ref{tbl:nlidb-example-1} shows two examples. In Example (a), we see that the natural language question mentions columns (a.k.a columns) of the database but the mentions and the column names are not exactly the same (e.g., ``director'' vs. ``directed by'', ``actor'' vs. ``star in'', etc.). In Example (b), we show that such differences could be quite significant, for example, ``population'' vs. ``how many people live in''. Furthermore, the question mentions ``Mayo'' but does not mention the column name ``county'' where ``Mayo'' appears in. Still, the NLI needs to derive the column name ``county'' from ``Mayo'' as it is needed in the translated SQL query. 


The challenge of NLIDB thus lies in the idiosyncrasy and complexity of a natural language: Column names in the schema and their mentions in the questions can have very different forms, and in some cases the mentions in the questions are missing, and need to be inferred from other parts of the question. As a result, an NLI developed for one particular database usually does work for another database.

In this paper, we explore the {\it latent semantic structure of natural language queries} against relational databases. As we can see in Figure 1, after we annotate the natural language queries (identifying mentions of database columns and values in the queries), the final annotated SQL queries for the two different questions against the two different databases are exactly the same. 
This means if we can separate out the idiosyncrasy of natural language, and focus on the underlying the semantics of relational queries, then it is possible to build a general purpose or transfer-learnable NLIDB. Experiments show that our method achieves $82\%$ query execution accuracy on WikiSQL~\cite{zhong2017seq2sql}, as well as zero-shot transfer-ability on OVERNIGHT~\cite{Wang2015BuildingAS}.

\eat{
In summary, our annotation mechanism automatically and explicitly provides a separation of \emph{semantics} and \emph{schema},
where previous works either rely on external, domain-specific knowledge bases to separate the semantics and the schema, or in the case of sequence to sequence translation learn semantics and schema jointly.
Our annotation mechanism is similar to slot filling (which is studied extensively in state tracking of dialogue systems) but with the additional mechanism of semantic and schema separation. Combined with a learned model, our method achieves $82\%$ query execution accuracy on WikiSQL~\cite{zhong2017seq2sql}, as well as zero-shot transfer-ability on OVERNIGHT~\cite{Wang2015BuildingAS}. 
}

\section{Overview of our Approach}

\begin{figure*}[thbp]	
\centering\includegraphics[width=.8\textwidth]{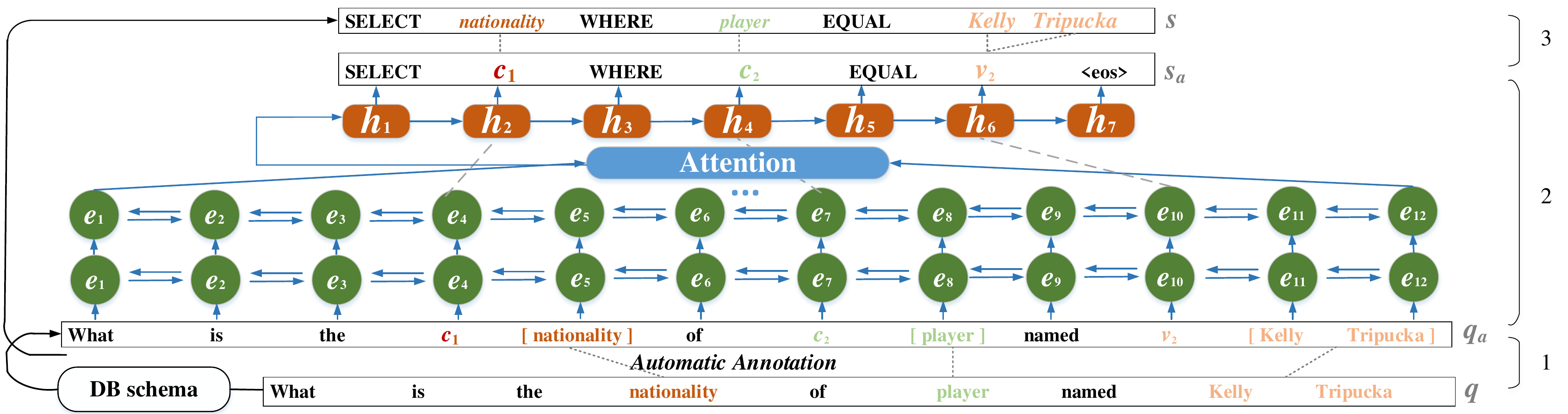}
\vspace{-5pt}
\caption{\label{fig:overview}
Framework overview.}
\vspace{-20pt}
\end{figure*}
Our goal is to separate out data specific components and focus on the latent semantic structure in a natural language question. The said data specific components include the schema of the data and the usage of natural language specific to the data schema, which we will describe in detail in the next section. 

Figure~\ref{fig:overview} shows the framework of our approach that consists of three stages: 

\begin{enumerate}
\item Convert a natural language question $q$ to its annotated form $q_a$;
\item Use a seq2seq model to translate $q_a$ to an annotated SQL $s_a$;
\item Convert the annotated SQL $s_a$ to a normal SQL $s$. 
\end{enumerate}

Figure 1 showed two examples of $(q, q_a, s_a, s)$.  In the annotation, we use  placeholder $c_i$ to denote the i-th column of a database table and $v_i$ to denote a value that is likely to belong to  the i-th column. 
For example, the term ``directed by'' in Figure~\ref{tbl:nlidb-example-1}(a) is annotated as $c_2$ since it is a mention of the 2nd column of the database table, and ``Jerzy Antczak'' is annotated as $v_2$ since it is a value of the second column. Note that a value that appears in a query may not appear in the database, thus, the annotation process must be able to annotate a term as a possible value of a database column. Also note that the process of converting an annotated SQL $s_a$ back to a normal SQL $s$ is deterministic. Thus, in the rest of the paper, we focus on step (1) and (2).

Our framework aims at separating natural language complexities out of relational query semantics. This allows us to focus on the remaining data agnostic and semantic part of the question with a seq2seq model. This simple idea is powerful because it reveals, for example, that the two different questions in 
Figure~\ref{tbl:nlidb-example-1} may have exactly the same structure \texttt{\small SELECT $c_1$ WHERE $c_2$ = $v_2 $ AND $c_3$ = $v_3$}). We consider the two natural language questions have the same ``latent semantic.'' In doing so,
our method mitigates the obstacles in transfer learning,
as a model trained on the first example could be applied to the second one due to shared annotated SQL query. 

\section{Annotation}
The task of annotating a natural language question is to detect mentions of database columns as well as database values in the question.

\subsection{Challenges}
Although some mentions of database columns and values can be detected exactly as they appear in a natural language question, 
in many cases, mention detection relies heavily on the context,
as illustrated in the following three examples of challenges:

\begin{enumerate}
\item {\it Who is the {best actress of year 2011}?} Here, ``best actress of year 2011'' mentions the database column ``best actor 2011''. Clearly, we cannot rely on exact string matching to detect mentions. 

\item {\it Which {film} {directed by} {Jerzy Antczak} did {Piotr Adamczyk} {star in}?  } (Example (a) in Figure~\ref{tbl:nlidb-example-1}).
Here, ``{Piotr Adamczyk}'' is an actor, but the question does not mention the database column ``actor'' explicitly. We need to infer the column from the question, the database schema, and the database statistics. 

\item {\it For which {player} his \textbf{rebounds} is 2 and \textbf{points} is 3.}''
Here, ``2'' and ``3'' could refer to either ``\textbf{rebounds}'' or ``\textbf{points}.'' The correct resolution depends on the syntax, or the context of the question.

\end{enumerate}

Next, we describe \textit{meta knowledge of a database}, which is the minimal knowledge about the underlying database we need in order to understand a natural language question against the database. 
Then we describe annotation as a two-stage process consisting of \textit{mention detection} and \textit{mention resolution}.
The first stage detects a set of many possible mentions of column and values, 
in which some mentions may be inconsistent with others since certain possibilities are mutually exclusive.
The second stage finds a maximum subset of these mentions that is consistent, which constitutes the output annotation.

\subsection{Meta Knowledge of a Database}\label{sec:schema}
We rely on the meta knowledge of  a database to understand questions posed against the database. The meta knowledge contains the following:
\begin{enumerate}
\item {Database schema $\mathcal{C}$.} The database schema is the definition of the columns of a database table. 
For example (a) in Figure~\ref{tbl:nlidb-example-1}, $\mathcal{C}$ = $\{$Nomination, Actor, Film Name, Director, Country$\}$.

\item {Database statistics $\mathcal{V}_c$ for each column $c \in \mathcal{C}$.}
Intuitively, a human database operator knows which database columns need to be involved for answering a particular question even if the name of the column is not explicitly mentioned in the question.
For example, in Figure~\ref{tbl:nlidb-example-1} (a), we should consult the ``Actor'' column for   ``Piotr Adamczyk''.
To approximate this mental process, 
we construct and leverage database statistics $\mathcal{V}_c$ that enables us to measure how likely a phrase is related to database column $c$ for all $c \in \mathcal{C}$.
For example, we may create a language model for a column. Specifically, we may use pre-trained word-embeddings to decide if a particular term belongs to a particular column (the idea is that if a term is related to a column, its embedding should be close to the word-embedding space of values in the column).

\item {Natural language expressions specific to a database, a database column, and values of the column $\mathcal{P}_c$.} 
We need to know how people talk about things embodied by the database. 
Ideally, if we have a general purpose ontology that tells us everything about how language is used to describe any entity and its features, 
we might not need this particular handcrafted component in the schema. 
But since such ontology does not exist, we consider this component requires minimal effort to enable us to support cross domain NLI. 
Specifically, for column $c$, we collect a set of phrases $\mathcal{P}_c$ that mention $c$. 
For example, 
for $c$ = {population} in Figure~\ref{tbl:nlidb-example-1} (b), we have $\mathcal{P}_c$ = $\{$population of $\langle$city$\rangle$, size of $\langle$city$\rangle$, how many people live in $\langle$city$\rangle$, ...$\}$.
Our approach provides a direct way to inject this minimal knowledge to our model. 

\end{enumerate}

\subsection{Mention Detection}
In \emph{Mention Detection}, we derive candidate mentions (of database columns and values) for terms in a question. Formally, we define a \emph{term} to be a continuous span of words. A term may {\it mention} a database column or value. For example, for the question {\it Which {film} {directed by} {Jerzy Antczak} did {Piotr Adamczyk} {star in}?  } in Figure~\ref{tbl:nlidb-example-1} (a), ``directed'' could be a mention of column ``Director'' and ``Piotr Adamczyk'' could be a mention of a value in either the ``Director'' column or the ``Actor'' column. Note that a term may have multiple candidate mentions, and the entire set of candidate mentions may not be consistent, and we rely on the next step, \emph{mention resolution}, to find a consistent set of mentions.



We treat mentions of database values and mentions of database columns differently. For mentions of columns, we have a set of known columns, and each of them can be mentioned in many different ways. For example, the term ``directed'' may be used to mention the ``Director'' column, and ``how many people live in ...'' may be used to mention the ``Population'' column. Our method, described below, combines syntax and semantics to deal with this scenario. For mentions of database values, however, we do not have a set of  ground-truth values. One may ask ``When was Joe Biden elected U.S. president?'' against a database of U.S. presidents, but clearly, ``Joe Biden'' is not in the database (at least not yet). Thus, for mentions of values $v$, we evaluate whether a term, in its exact literal form in the question (denoted as $w_v$), is likely to be a value of a column of the database.


We now describe mention detection for columns. For a term, defined as a span of words $w=[w_a, w_{a+1}, \dots, w_{b}]$ in a question, we want to know if it is a mention of a column $c$. 
We do the following:

\begin{itemize}
	\item
We consider $w$ a possible mention of column $c$ only if the $w$ covers $c$ as a string \emph{effectively and efficiently}. 
Formally, $w$ covers $c$ if 
(1) There does not exist a larger span $w'$ that contains $w$ (e.g. $w'=[w_{a'}, w_{a'+1}, \dots, w_{b'}]$ s.t.~ $a'\leq a, b \leq b', |a-a'| + |b'-b| \geq 1$) and covers more words of $c$ than $w$;
(2) There does not exist a smaller span $w'$ that is contained within $w$(e.g. $w'=[w_{a'}, w_{a'+1}, \dots, w_{b'}]$ s.t.~ $a\leq a', b' \leq b, |a-a'| + |b'-b| \geq 1$) and covers the same number of words of $c$ as $w$.
The computation of number of words covered is done by counting the number of pairs ($x$, $y$) where $x \in w$ and $y \in c$ such that $x$ and $y$ are \emph{close}.
The measure of \emph{closeness} could be either of the following:
\begin{itemize}
\item \emph{Edit Distance}. We denote $F_{\mathtt{ed}}(x,y) = {D(x,y)}$ $/$ ${ \max\left( |x| , |y| \right)}$  where $D$ is character-level edit distance,
and define $x$ and $y$ are \emph{close} if $F_{\mathtt{ed}}(x,y)<\tau_{\mathtt{ed}}$.
\item \emph{Word-Embedding Space Distance}. We denote $F_{\mathtt{sem}}(x,y) $ $ = 0.5 \cdot( 1 - \mathtt{Cosine}\left( \mathcal{W}(x), \mathcal{W}(y) \right) $,
and define $x$ and $y$ are \emph{close} if $F_{\mathtt{sem}}(x,y)<\tau_{\mathtt{sim}}$.

\end{itemize}

\item
We use meta database knowledge $\mathcal{P}_c$ directly for mention detection. 
For example, in Figure~\ref{tbl:nlidb-example-1}(b),
the phrase ``how many people live in'' is detected as a mention for column ``Population''.
\end{itemize}

\eat{
The measure of \emph{closeness} between of two words $x$ and $y$ uses edit distance and word-embedding space distance.
In detail, we denote $F_{\mathtt{ed}}(x,y) = {D(x,y)} / { \max\left( |x| , |y| \right)}$  where $D$ is character-level edit distance,
and $F_{\mathtt{sem}}(x,y)= 0.5 \cdot( 1 - \mathtt{Cosine}\left( \mathcal{W}(x), \mathcal{W}(y) \right) $.
Both metrics have range of $[0, 1]$ and smaller value indicating more closeness.
We define  $x$ and $y$ are \emph{close} if $F_{\mathtt{ed}}(x,y)<\tau_{\mathtt{ed}}$ or $F_{\mathtt{sem}}(x,y)<\tau_{\mathtt{sim}}$.
}

\eat{
Finally, for all table columns that are not paired by with any value but with \emph{table-related natural language expressions $P_c$} available, 
This process proposes $c$ and its mention $\mathcal{P}_c$.
For example in Figure~\ref{tbl:nlidb-example-1}(b),
column ``population'' is detected by sub-phrase ``how many people live in''.
}

Taking question $q$ = ``{ Who is the best actress of year 2011?}'' and column name $c$ = ``{best actor 2011}'' as an example. We detect $w$ = ``best actress of year 2011'' as a mention of $c$, as ``best'', ``actress'' and ``2011'' in $w$ have words covered in $c$, namely ``best'', ``actress'', ``2011''. Note that the closeness of ``actress' and ``actor'' is assessed by $F_{\mathtt{sem}}$.
Moreover, $w$ is neither  ``the best actress of year 2011?'' (violating (1) above) nor ``best actress of year'' (violating (2) above).

\begin{figure}[h!]
	\vspace{-5pt}
	\centering
	\includegraphics[width=0.35\textwidth]{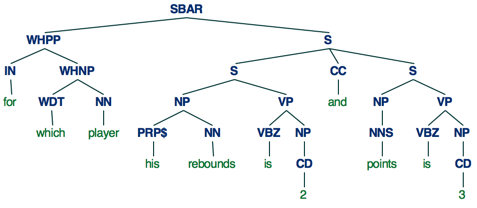}
	\vspace{-10pt}
	\caption{Use lowest common ancestor (LCA) in constituency tree to find correct pairing of column and value.
		Here ``rebounds'' is paired with ``2'' and ``points'' with ``3''.}
	\label{fig:tree}
	\vspace{-5pt}
\end{figure}

\subsection{Mention Resolution}

There could be many candidate mentions of columns and values for many terms. The goal of mention resolution is to figure out globally, what is the most likely subset of mentions that are consistent.

First, we use syntax or context to reduce the number of candidate mentions. For example, in question ``{For which {player} his \textit{rebounds} is 2 and \textit{points} is 3?}'' the values
``2'' and ``3'' could be mentions of either ``\textit{rebounds}'' or ``\textit{points}''. Both are valid unless we take into consideration the syntax and context of the question. 
We observe that in the question's constituency tree, paired columns and values are structurally close to each other. Therefore, we use the depth of Lowest Common Ancestor (LCA) in the question's constituency tree as a measure of structural closeness.
More specifically, for  value $v$ and column $c$, and their mentions $w_v$ and $w_c$ in the question respectively,
the structure closeness is measured by $C(v, c) = \max_{x^v \in w_v , x^c \in w_c} \LCA_{\mathrm{depth}} (x^v, x^c)$ where $x^v, x^c$ are words in $w_v, w_c$ respectively.
We  only consider  matches $(v, f)$ that have best structure closeness, e.g.~ when $C(v, c) = \max_{c\in \mathcal{C}} C(v, c)$.

Second, we use a global metric to further reduce inconsistency. Note that the first step does not solve all the problems. For example, it is possible that multiple column candidates have the same maximal LCA to a given value in the question.
We formalize this task as a graph matching problem: 
Consider a bipartite graph $G=(V,C,E)$ where vertices $V$ represent all mentions of values, vertices $C$  all mentions of columns, 
and edges $E$ contain edges between two vertices $(v\in V, c \in C)$ if $(v, c)$ is considered as a possible match as described above.
We then find a Maximum Bipartite Matching (MBM)~\cite{kuhn1955hungarian} in $G$ as a proxy to the mention resolution,
since finding MBM is  equivalent to finding a maximum subset of matchings that are compatible,
where each edge $(v,c)$ in MBM constitutes a matching of value and column in the output annotation. After finding MBM, we add all values and columns whose corresponding vertex is without any edge in $G$  to the annotation, since adding them would not disturb any existing matching.
Finally, all columns and values are deterministically given an index ($c_1, c_2, \dots \quad v_1, v_2, \dots$) where (1) matching column and value share the same index, and (2) indices are ordered according to the earliest mention of indices' corresponding column or value in the question.

\section{Sequence to Sequence Translation}\label{sec:proposed_model}

We denote an input natural language \emph{question} as $q$ = $(q_1, q_2, ... ,q_m)$,
\textit{annotated question} as $q_a$ = $(q_a^1, q_a^2, ... , q_a^m)$,
and the corresponding \textit{annotated SQL query} as $s_a$ = $(s_a^1,s_a^2,...,s_a^n)$.
The result of the aforementioned annotation process on a natural language question $q$ is represented as its annotated form $q_a$,
which is the input to a seq2seq model that translates $q_a$ to an annotated SQL $s_a$.
In this section we discuss the representation of the annotated input, 
followed by a \emph{Sequence Translation Model}  that produces $s_a$ for a given $q_a$.

\subsection{Representation of Annotated Input}

There are many options in representation. For example, in Figure~\ref{tbl:nlidb-example-1}(a), ``directed by'' is annotated as the mention of $c_2$. We can either substitute ``directed by'' by $c_2$ or insert $c_2$ after ``directed by'' in the question. We explore difference annotation encoding methods and propose our unique annotation encoding to separate schema from natural language questions without loss of schema information. We will show in the Experiment Section that the representation affects the schema separation performance dramatically.

\begin{figure}[h]
\vspace{-5pt}
	\begin{flushleft} 
		\scriptsize  \textbf{\textit{Question}}  \quad
		\texttt{What position did the player LeBron James play?}
		\scriptsize  \textbf{\textit{Annotated Question: Symbol Substitution}} \quad
		\texttt{What $c_1$ did the $c_2$ $v_2$}\\
		\scriptsize \textbf{\textit{Annotated Question: Symbol Appending}} \quad
		\texttt{What $c_1$\colorbox{green!20}{[position]} did the $c_2$\colorbox{green!20}{[player]} $v_2$\colorbox{green!20}{play}?}
	\end{flushleft}
	\vspace{-10pt}
	\caption{An example with two format of annotated form.}
	\label{tbl:annotation}
	\normalsize
	\vspace{-10pt}
\end{figure}

\subsubsection{Include annotations as additional symbols}
Intuitively, the annotation should enable schema separation that strips off schema-specific information from natural language questions or substitutes schema-specific information with symbols. However, replacing the mentions of columns with unified symbols like $c_i$ or $v_i$ discards the diverse semantic meaning of the column texts. Therefore we propose to append the symbols into stacks rather than substituting them to leverage the semantics of column texts.
This proposed approach is referred to as ``column stack'' in Experiment Section. 
Figure~\ref{tbl:annotation} shows the difference of the two approaches and highlights that the proposed approach provides more information to the downstream sequence model.


\begin{figure}[h!]
\vspace{-5pt}
	\small 
	\begin{flushleft} 
		\scriptsize  \textbf{\textit{Annotated Question:}}\quad \texttt{When $v_1$ [Piotr Adamczy] was nominated as $c_1$ [Best Actor in a Leading Role] ?}\\
		\scriptsize  \textbf{\textit{Annotated SQL}} \quad \texttt{SELECT \colorbox{green!20}{Nomination Date} WHERE $c_1$ = $v_1$}\\
		\vspace{1em}
		\scriptsize  \textbf{\textit{Annotated Question  with table header encoding:}}\quad \texttt{When $v_1$ [Piotr Adamczy] was nominated as $g_1$ [Best Actor in a Leading Role] ? \quad | \quad $g_1$ [Nomination] $g_2$ [Actor] $g_3$ [Film Name] $g_4$ [Director] $g_5$ {[Nomination Date]}}\\
		\scriptsize  \textbf{\textit{Annotated SQL with table header encoding}} \quad
		\texttt{SELECT \colorbox{green!20}{$g_5$} WHERE $c_1$ = $v_1$}
	\end{flushleft}
	\vspace{-10pt}
	\caption{An example showing approaches with and without table header encoding.}
	\label{tbl:table-header}
	\normalsize
	\vspace{-10pt}
\end{figure}

\subsubsection{Table Header}

When a column in SQL is not mentioned in the question, we can only rely on the sequence model to infer the column.
For the example in Figure~\ref{tbl:table-header}
, ``Nomination Date'' is not explicitly mentioned and needs to be inferred by the model. 
However, most of the columns  (e.g., ``Nomination Date'') consist of multiple tokens, which are hard to be inferred correctly by a sequence model.
To encourage the correct inference of multi-token columns,
we append all headers $g \in \mathcal{C}$ in the end of the annotated question,
so that even if a column is not mentioned in and cannot be inferred only from the question, 
it could still be referred to as $g_i$ by the sequence model.

Figure~\ref{tbl:table-header} shows an example where 
``\texttt{$g_1$ [Nomination] $g_2$ [Actor] $g_3$ [Film Name] $g_4$ [Director] $g_5$ {[Nomination Date]}}''
is appended to the annotated question.
and thus simplifies the annotated SQL to the form of ``\texttt{\small SELECT {$g_5$} WHERE $c_1$ = $v_1$}''. 
Besides referring to unmentioned headers,
our strategy also contributes to a much smaller output vocabulary and thus makes it easier to train a sequence model.

\subsection{Sequence Translation Model}
We train a seq2seq model to estimate $p(s_a|q_a)$, which
captures the conditional probability of $p(s_a|q_a)$ = $\prod^n_{j=1}p(s_a^j|q_a,s_a^{1:j-1})$ with encoder and decoder.

\textbf{Encoder}
is implemented as stacked bi-directional multi-layer recurrent neural network (RNN) with using Gated Recurrent Unit (GRU)~\cite{cho2014learning}. 
To keep the dimension consistent, we add an affine transformation before
each layer of RNN, which is defined as
${y_i}^{(l)} = {W_0}^{(l)} x^{(l)}_i + {b_0}^{(l)}$,
where ${x_i}^{(l)}$ is the input of the $l$-th layer in the $i$-th position.
${W_0}^{(l)}$ and ${b_0}^{(l)}$ are model parameters.
The hidden state of the forward RNN and backward RNN are computed as:
\footnotesize
\begin{align*}
\overrightarrow{h_i}^{(l)} &= \mathrm{GRU}_{\rightarrow}({y_i}^{(l)},{\overrightarrow{h_{i-1}^{(l)}}})\\
\overleftarrow{h_i}^{(l)} &= \mathrm{GRU}_{\leftarrow}({y_i}^{(l)},{\overleftarrow{h_{i-1}^{(l)}}})
\end{align*}
\normalsize

We concatenate forward state vector and backward state vector as ${h_i}^{(l)} = [{\overrightarrow{h_i}}^{(l)}, {\overleftarrow{h_i}}^{(l)}]$, $i=1,2,...,m$.
The input of each layer is computed as
${x_i}^{(1)} = \phi(q_a^i) $   ($\phi$ is the word embedding lookup function) $ {x_i}^{(l+1)} = {h_i}^{(l)}$.


\textbf{Decoder} is an one-layer \emph{attentive} RNN with \emph{copy mechanism}.
We use Bahdanau's attention~\citep{bahdanau2014neural} as follows:
At each time step $i$ in the \emph{decoder}, the decoding step is defined as:
\footnotesize
\begin{align*} 
d_0 &= \Tanh(W_1 [{\overrightarrow{h_m^{(l)}}}, {\overleftarrow{h_1^{(l)}}}])\\
d_{i} &= \mathrm{GRU}([\phi(s_a^{i-1}),\beta_{i-1}],d_{i-1}) \\ 
e_{ij} &= v^T \Tanh(W_2 {h_j^{(l)}} + W_3 d_i) \\
\alpha_{ij} &= {e_{ij}} \large/ {\sum_{j'} e_{ij'}} \\
\beta_i &= \sum^m_{j=1} \alpha_{ij}  {h_j}^{(l)}
\end{align*}
\normalsize
where $W_1$, $W_2$, $W_3$, $v$, $U$ are model parameters,
$(d_1,d_2,...,d_n)$ the hidden states of the decoder RNN
and $j$ the index enumerating all positions in the \emph{encoder}.
In NLIDB, annotations in the output SQL often correspond directly from the input natural language question, 
To encourage the model choose tokens that appears in the input.
We introduce copy mechanism that samples output token $s_a^t$ as
\footnotesize
\begin{align*} 
p(s_a^i|q_a,s_a^{1:i-1}) &\propto \exp(U[d_i,\beta_i]) + M_i\\
M_i[s_a^j] &= \exp(e_{ij})
\end{align*}
\normalsize
Note that this is different from the vanilla copy mechanism where the output is sampled through softmax over entire word vocabulary as $p(s_a^i|q_a,s_a^{1:i-1}) \propto exp(U[d_i,\beta_i])$.

\section{Related Work}

\begin{table*}[t!]
	\centering
	\resizebox{0.80\textwidth}{!}{%
		\begin{tabular}{clcccccc}
			\toprule
			\multicolumn{2}{c}{Model} &  \multicolumn{3}{c}{Dev} & \multicolumn{3}{c}{Test}\\
			&  &  Acc$_{\mathrm{lf}}$  &  Acc$_{\mathrm{qm}}$   & Acc$_{\mathrm{ex}}$  & Acc$_{\mathrm{lf}}$ & Acc$_{\mathrm{qm}}$ & Acc$_{\mathrm{ex}}$ \\
			\midrule
			& Seq2SQL~\cite{zhong2017seq2sql} & 52.5\% & 53.5\% & 62.1\% & 50.8\% & 51.6\% & 60.4\%\\
			Previous & SQLNet~\cite{sqlnet}   &  - & 63.2\%  & 69.8\%  & - & 61.3\% & 68.0\%\\
			Method & PT-MAML~\cite{huang2018natural} & 63.1\% & - & 68.3\% & 62.8\% & - & 68.0\% \\ 
			& TypeSQL~\cite{typesql}* & -  & 68.0\%  & 74.5\% &  - & 66.7\% & 73.5\% \\
			& Coarse2Fin~\cite{coarse2fine} & - & - & - & 71.7\% & - & 78.5\% \\ 
			\midrule
			Annotation & Annotated Seq2seq (Ours)   &   $\mathbf{72.0\%}$ & $\mathbf{72.1\%}$ & $\mathbf{82.1\%}$  & $\mathbf{72.0\%}$ & $\mathbf{72.1\%}$ & $\mathbf{82.2\%}$ \\
			Based 
			& \quad\quad -- 1 GRU Layer & 71.6\% & 71.7\% & 81.2\% &  71.6\% & 71.7\% & 81.6\% \\
			& \quad\quad -- Copy Mechanism &  71.6\% & 71.6\% & 81.6\% & 71.6\% & 71.6\% & 81.3\% \\
			& \quad\quad -- Column Stack  & 72.1\% & 72.2\% & 81.6\% & 71.8\% & 71.8\% & 81.1\% \\ 
			& \quad\quad -- Table Header Encoding & 71.9\% & 71.9\% & 79.6\%& 71.6\%& 71.7\%& 79.6\% \\ 
			& \quad\quad -- seq2seq + Transformer  &  64.7 \%& 64.7\%  & 78.8\%  & 65.3\% & 65.4\% & 79.2\%\\
			\bottomrule
		\end{tabular}
	}
	\caption{Comparison of models. 
		$\mathrm{lf}$, $\mathrm{qm}$, $\mathrm{ex}$ represent logical forms,   exact query match, and  query execution accuracy respectively. 
		Performance on the first block are copied from the corresponding papers. ``--'' and ``+'' mean removing or adding one component from our best approach respectively for ablation.\\
		* We report performance without using extra knowledge base for a fair comparison}
	\label{table:expr_comparison}
	\vspace{-15pt}
\end{table*}


\paragraph{Natural Language Interface to Database (NLIDB)}
Natural Language Interface to Database aims at providing an interactive bridge between humans and machines,
where human users can issue a question in natural language text,
which would then be translated to a structured query that can be executed by a  database engine.
\citet{androu1995natural} first explores this task with concrete examples defining this problem and
highlights the separation between linguistic and database-derived information.
Later \citet{popescu2003towards} proposes to identify questions whose answers are tractable solely from the database,
and \citet{giordani2012translating} incorporates tree kernels in ranking candidate queries.
Many recent advances can be categorized into two groups.
The first group uses semantic parsing~\cite{Wang2015BuildingAS,PasupatL15,Jia2016DataRF}
as well as some extensions that support cross-domain semantic parsing~\cite{herzig2017neural,su2017cross}. However, due to the idiosyncrasy and complexity of natural languages, most works in this group are confined in narrow domains.
The other group relies on neural based methods where sequence to sequence models are leveraged to translate input questions to SQL queries,
optional combined with the help of user feedback~\cite{iyer2017learning} and reinforcement learning~\cite{zhong2017seq2sql}.

\paragraph{Slot Filling in Dialogue System}
Dialogue system aims at communicating with a user in a session with multiple turns of dialogs,
where \emph{state}, or conceptually what the session is talking about, needs to be tracked for dialogue system to archive good performance \cite{young2010hidden}.
Commonly the dialogue system identifies and tracks entities that appear across turns as slots in a process called \emph{slot filling}.
These slots and entities that can fill in these slots are usually specific to the domain that the dialogue system is focusing on.
For example, slots can be food, airport or city names, and therefore are from a pre-defined, externally crafted list of possible values.
Recently, neural based approaches have been proposed for tracking state:
\citet{henderson2014word} proposes a simple recurrent network that predicts the probability of each word in dialog being one of pre-defined slots,
which is extended by \citet{mrkvsic2015multi}, a hierarchical model that can handle cases where entities can be from one of multiple domains.
To specifically provide better tracking for ranking slot values in dialog,
Belief Tracker~\cite{mrkvsic2017neural} sums up separating representations of system output, user feedback and candidate slot values.
To further improve the performance, \citet{wen2017network} considers a policy network that arbitrates the outputs from multiple models,
including the aforementioned belief tracker, a sequence model that encodes user input, and a generation network that produces system output.

Closest to our proposed work is~\cite{sqlnet}, which  employs a sketch-based approach that represents
an SQL as a template with slots, and the model predicts values from a limited candidate set to be filled in each slot. This is different from our work that focuses on annotation and does not restrict SQL to a particular template-based form.
Another close work is TypeSQL~\cite{typesql}
that enriches the inference of columns and values using domain-specific knowledge-based model that searches five types of entities on Freebase, an extra large database,
which is contrast to our work which does not rely on extra database knowledge.

\paragraph{Sequence-to-sequence Generation model} 
Sequence to Sequence (referred to as seq2seq in the rest of the paper) learning~\cite{sutskever2014sequence}
has led to many advances in neural semantic parsing models.
Notable advances in sequence learning include
attention mechanism \cite{bahdanau2014neural} and pointer network \cite{vinyals2015pointer}
that boost performance for sequence learning and enable it to handle long sequence and rare words.
They have seen successful applications on
language model~\cite{merity2016pointer}, text summarization~\cite{gu2016incorporating},
 text understanding~\cite{wen2017network}, and neural computing~\cite{graves2016hybrid}.
Our model also benefits from these techniques since
our model needs to see both information packed in long sequence and rare words that only appear in few tables.


\section{Experiments and Analysis}

We conduct experiments\footnote{The data and code publicly is available at \url{https://drive.google.com/open?id=1YAXogJ8H5iRLTsDpko5xa6ZIX60dal6-}}
on two scenarios: (1) in-domain scenario of NLIDB trained and evaluated on WikiSQL dataset~\cite{zhong2017seq2sql},
and (2) cross-domain scenario where we evaluate our model (trained on WikiSQL)'s  transferring accuracy on OVERNIGHT dataset~\citep{wang-yang:2015:EMNLP} without extra training in the target domain.

We use three metrics for evaluating the query synthesis accuracy:
\textbf{(1) Logical-form accuracy}. We compare the synthesized SQL query against the ground truth for whether they agree token-by-token in their logical form,
as proposed in~\cite{zhong2017seq2sql}.
\textbf{(2) Query-match accuracy}. Like logical-form accuracy, 
except that we convert both synthesized SQL query and the ground truth into canonical representations before comparison.
This metric can eliminate false negatives cases,
such as two semantically identical SQL queries being different only in the literal ordering of condition clauses.
\textbf{(3) Execution accuracy}. We execute both the synthesized query and the ground truth query
and compare whether the results agree,
as proposed in~\cite{zhong2017seq2sql}.

\subsection{WikiSQL}
We train and evaluate our NLIDB model on WikiSQL,
which contains $87673$ records of natural language questions, SQL queries, and $26521$ database tables.
Since tables are not shared among the training / validating / testing splitting provided in the dataset, 
models evaluated on it are supposed to generalize to new questions and database schema. 

\eat{
\paragraph{Automatic Annotation Coverage}

\begin{table}[t!]
\small
\centering
\resizebox{0.27\textwidth}{!}{%
\begin{tabular}{ c c c c c c c}
\toprule
Match   & MCBM & Column Inference \\
\midrule
Train  & 26.89\% & 27.69\% \\
\midrule
Dev &  25.87\% & 27.69\%\\
\midrule
Test & 26.50\% & 28.17\%\\
\bottomrule
\end{tabular}
}

\caption{Automatic annotation component coverage. We report the coverage of (MCBM) Maximum Continuous Bipartite Matching and Column Inference, respectively, on training / validating / testing split provided by WikiSQL dataset.}
\label{table:match_coverage}
\normalsize
\vspace{-5pt}
\end{table}

To evaluate how each component of automatic annotation performs on 
WikiSQL dataset,
we report the percentage of records such that one or more sub-phrases of the natural language question are identifies as mentions of column names by each component.
As shown in Table~\ref{table:match_coverage}, we report the coverage of Maximum Continuous Bipartite Matching (MCBM) and Column Inference.
The coverages of all components do not sum to $100$\% because each record could possibly use more than one strategies.
The experimental results show that
MCBM provides more than $25$\% coverage, and 
approximately $28$\% implicit columns could be inferred.
Such benefit enables the model to explicitly capture information that could only be memorized implicitly.
}

\paragraph{Training Details} 
The annotation process uses $\tau_{\mathtt{ed}}=0.5$ and $\tau_{\mathtt{sim}}=0.15$.  
For the sequence model, we use two layers of stacked GRU with hidden size $200$ for encoder, and $400$ for decoder.
The input and output layers of both encoder and decoder share the tied embedding weights.
We initialize the embedding weights with pre-trained Glove embedding with dimension $D = 300$,
and embeddings for tokens not covered by GloVe with random vector. 
Symbols introduced by annotation (such as $c_1$) are also treated as tokens,
each of them being represented by the concatenation of the embedding of annotation type ($c$ or $v$) and index.
Also, the embedding of annotation type and index are randomly initialized with $D' = 150$ so the concatenations has the same dimension as $D=300$.
In training we use gradient clipping with threshold $5.0$,
and in inference we use beam search with width $5$.

\paragraph{Evaluation}

\begin{table*}[t!]
	\centering \small
	\resizebox{0.80\textwidth}{!}{%
		\begin{tabular}{l*6c}
			\toprule
			\multicolumn{1}{c}{Domain}&Basketball&Calendar&Housing&Recipes&Restaurants&Overall\\
			\midrule
			\multicolumn{1}{l}{\qquad Pruning Rate}  & 41.33\%  &  59.80\% & 52.44\% & 79.33\% & 38.33\% & 49.44\%\\
			\multicolumn{1}{l}{\qquad$\#$ of records} & 1130 & 320 & 429 & 203 & 983 & 3065 \\
			\midrule
			\qquad Transfer Acc$_{qm}$ & 54.60\% & 80.31\% & 48.48\% & 75.86\% & 82.81\% & 66.88\%\\
			\quad\qquad -- Column Stack & 48.23\% & 61.25\% & 28.21\% & 75.37\% & 36.83\% & 44.93\% \\
			\quad\qquad -- Table Header Encoding & 58.14\% & 73.44\% & 51.98\% & 82.27\% & 84.74\% & 69.0\% \\
			\quad\qquad -- 1 GRU layer & 48.67\% & 74.06\% & 54.78\% & 80.30\% & 84.74\% & 65.84\% \\
			\quad\qquad -- Copy Mechanism & 59.56\% & 80.31\% & 51.52\% & 81.77\% & 79.15\% & 68.35\% \\
			\bottomrule
		\end{tabular}
	}
    \vspace{-5pt}
	\caption{OVERNIGHT transfer accuracy. ``--'' means removing one component from our best approach respectively for ablation.}
	\label{table:expr_transfer}
\vspace{-15pt}
\end{table*}

We compare our method with previous methods
through three aforementioned metrics: accuracies in terms of logical form exact match, exact query match, and the results of query execution. 
As shown in Table~\ref{table:expr_comparison},
our result outperforms these previous methods,
including the state-of-the-art Coarse2Fin~\cite{coarse2fine} by $0.3\%$ for exact query match, and $3.7\%$ for query execution.
This demonstrates that our method 
enables high transfer-ability to unseen tables,
since in WikiSQL database tables are not shared among training and testing splittings.

We note that TypeSQL achieves 
high accuracy in the "content-sensitive" setting where it queries Freebase when handling natural language questions in training as well as in inferencing. TypeSQL is thus limited because it cannot easily generalize to content that is not covered by Freebase.
Since our focus is automated process with generalization, 
we did not take this setting into consideration.

\paragraph{Ablation}\label{sec:ablation}
In Table~\ref{table:expr_comparison}, we conduct ablation analysis to demonstrate the contribution to performance from different components of our model.
Removing each component of our method leads to a decreases in performance:
The removal of (1) a layer in encoder (e.g. using only one layer of GRU), (2) copy mechanism,
(3) column stack (e.g. using column substitution ), and (4) encoding of table header,
each respectively decrease performance on testing set. 

Since in our framework the annotation and sequence modeling are separate processes,
we also test our annotation method combined with the transformer model~\footnote{We use pre-trained GloVe embedding, 
transformer implementation from  tensor2tensor
 \url{https://github.com/tensorflow/tensor2tensor}
with \texttt{hiddien\_size = 300, num\_heads = 6}},
an alternative and state-of-the-art architecture for sequence modeling such as machine translation.
With the same annotation,
the transformer model shows worse performance. 
We hypothesis that this is due to
NLIDB task being different from translation tasks:
NLIDB has huge difference between vocabulary sizes in source space and target space, and only outputs the symbols while learning original text implicitly.

\subsection{Cross-domain transfer-ability}

For cross-domain evaluation, we evaluate the transfer-ability of our model that is trained on one domain (WikiSQL in our case) and tested on
other domains to assess its transfer-ability.
This task is challenging since the model is required to model domains not seen before.

\paragraph{Dataset}
We use OVERNIGHT dataset~\citep{wang-yang:2015:EMNLP},
consisting of pairs of natural language query and dataset-specific logical form,
as target domains.
In particular, 
the query and logical form pairs are originally categorized into eight sub-domains,
of which we use five of them (\textsc{Basketball}, \textsc{Calendar}, \textsc{Housing}, \textsc{Recipes}, and \textsc{Restaurants}) 
where the logical form can be feasibly converted to SQL for evaluation.
We then annotate these five sub-domains of OVERNIGHT dataset.

\paragraph{Evaluation}
We evaluate the transfer-ability of our model that is trained on WikiSQL on five sub-domains of OVERNIGHT dataset.
Since OVERNIGHT SQL sketch is highly variant and different from WikiSQL sketch. We make a reasonable assumption that only SQL sketch compatible with WikiSQL are considered in the transfer-ability evaluation. 
Table~\ref{table:expr_transfer} presents our transfer-ability performance,
where \textit{Pruning Rate} represents the percentage of sketch non-compatible records in each category, 
and \textit{number of records}
represents the number of sketch compatible records in each category, 
which is the size of our evaluation set.
Transfer accuracy is calculated over sketch compatible records and based on recovered SQL.
It is shown that our model performs high transfer-ability with zero-shot learning on OVERNIGHT dataset.

\paragraph{Ablation}
We conduct ablation analysis to demonstrate the transfer-ability contributed by our annotation strategies as well as different components of our model as shown in Table~\ref{table:expr_transfer}.
When we remove column stack, the model not only 
performs worse for WikiSQL (see Table~\ref{table:expr_comparison}), but also transfers poorly to OVERNIGHT. The overall query match accuracy is reduced by 22\%.
In contrast to the results shown in Table~\ref{table:expr_comparison}, encoding table headers hurts transfer ability.
The reason is that,
in OVERNIGHT dataset, only five sub-domains are considered, so we use pre-collected paraphrase set to annotate columns.
Most of the records are fully annotated, and encoding table headers does not enhance annotation coverage. On the contrary, it feeds the model redundant information which hurts the overall accuracy.

We also conduct ablation analysis to demonstrate the transfer-ability contributed by different component of our model.
As shown in Table~\ref{table:expr_transfer},
the removal of 1 GRU layer results in 1\% overall accuracy decrease. The model using only one layer of GRU is not capable of capturing all the conditional probabilities, thus causing lower transfer-ability. 

Counter-intuitively, the model without copy mechanism has 1.5\% higher transfer accuracy than the original model. Copy mechanism contributes to the test accuracy on Wiki\-SQL but hurts the transfer-ability on OVERNIGHT. 
We hypothesis that transfer-ability 
is not fully correlated with the accuracy of the original model, it also
depends on schema extraction and how the model generalizes to another domain.

\eat{
\paragraph{Case Study}

\begin{table*}[!htbp]
\centering \small
\begin{tabular}{*7c}
\toprule
Domain&Basketball&Calendar&Housing&Recipes&Restaurants\\
\midrule
Anno  & 0.05\%  & 0.05\% & \% & \% & \% \\
\bottomrule
\end{tabular}
\caption{OVERNIGHT fall-out domain transfer accuracy w/o annotation}
\label{table:expr_transfer}
\end{table*}

\eat{
\begin{figure}[h]
\begin{flushleft} \small \textbf{Question:} \end{flushleft}
\begin{center} \small
Show me all meetings Alice has attended at the Greenberg Cafe.
\end{center}
\begin{flushleft}\small \textbf{Baseline SQL Query:} \end{flushleft}
\begin{center} \small \texttt{ SELECT meeting where \colorbox{red!30}{location} equal Alice and location equal Greenberg Cafe} \end{center}
\begin{flushleft} \small \textbf{Our SQL Query:} \end{flushleft}
\begin{center} \small \texttt{ SELECT meeting where \colorbox{green!20}{attendee} equal Alice and location equal Greenberg Cafe} \end{center}
\caption{Example of correctly inferred aggregate.}
\label{tbl:nlidb-example-rightinfer}
\end{figure}

We reproduce the model proposed in~\cite{su2017cross} with SQL queries as output, denoted as \textit{Su\&Yan}.
First of all, we present some cases where SQL queries are falsely inferred by \textit{Su\&Yan} but correctly generated by our method (shown in Figure~\ref{tbl:nlidb-example-rightinfer}). 
In \textit{Su\&Yan} method, ``location'' is inferred as the column name in the first sub-clause.
}

\begin{figure}[h]

\begin{flushleft} \small \textbf{Annotated Question:} \end{flushleft}
\begin{center} \small
How many \underline{blocks} \texttt{$c_1$} did 
\underline{player} \texttt{$c_2$} \underline{Kobe Bryant} \texttt{$v_2$} make during seasons
where he made \underline{3} \texttt{$v_3$} \underline{assists} \texttt{$c_3$}?
\end{center}
\begin{flushleft}\small \textbf{SQL Query:} \end{flushleft}
\begin{center} \small \texttt{ SELECT $c_1$ WHERE $c_2$ = $v_2$ AND $c_3$ = $v_3$} \end{center}
\begin{flushleft} \small \textbf{Inferred SQL Query:} \end{flushleft}
\begin{center} \small \texttt{COUNT SELECT $c_1$ WHERE $c_2$ = $v_2$ AND $c_3$ = $v_3$} \end{center}
\vspace{-10pt}
\caption{Example of wrongly inferred aggregate.}
\label{tbl:nlidb-example-wronginfer}
\vspace{-10pt}
\end{figure}

We show some wrongly inferred examples.
Some of wrong inferences are due to aggregate
as shown in Figure~\ref{tbl:nlidb-example-wronginfer}.
The reason is that, \underline{blocks} is identified as "number of blocks" column,
and the annotated sentence starts from "How many $c_1$". When the model sees "how many"
before $c_1$, it adds "COUNT" to the output naturally.
In theory, this type of error is due to annotation glitch. The model is functioning well, and we will improve our annotation to avoid such errors.
Figure~\ref{tbl:nlidb-example-wronginfer-1} shows an example of
wrongly inferred operator. In this example, "after" indicates ">" operator. However, the model has not seen such situation in the training, so ">" operator can not be inferred correctly. Even schema is separated from natural language,
the model can not infer correctly if the language is never seen during training.
In summary, our method separates schema from natural language, which contributes to
a high transfer-ability NLIDB model.

\begin{figure}[h]
\begin{flushleft} \small \textbf{Annotated Question:} \end{flushleft}
\begin{center} 
Show me \underline{meetings} \texttt{$c_1$} after \underline{date} \texttt{$c_2$} \underline{Jan 2nd} \texttt{$v_2$}
\end{center}
\begin{flushleft} \small \textbf{SQL Query:} 
\quad  \small \texttt{ SELECT $c_1$ WHERE $c_2$ > $v_2$ } \end{flushleft}
\begin{flushleft} \small \textbf{Inferred SQL Query:} 
\quad \small \texttt{SELECT $c_1$ WHERE $c_2$ = $v_2$} \end{flushleft}
\vspace{-10pt}
\caption{Example of wrongly inferred operator.}
\label{tbl:nlidb-example-wronginfer-1}
\vspace{-10pt}
\end{figure}
}

\section{Conclusion}
In this work, we propose an NLIDB as a general paradigm to convert natural language queries to SQL queries for any database. 
The main contribution of our work is to separate meta data from the data itself and learn, transfer, and accumulate knowledge of natural language and domain-specific knowledge separately. Our extensive experimental analysis demonstrates the advantage of our approach over state-of-the-art approaches in standard datasets.

\small

\bibliography{bibfile}
\bibliographystyle{aaai}

\end{document}